\documentclass{article}

\usepackage{PRIMEarxiv}

\usepackage[utf8]{inputenc} 
\usepackage[T1]{fontenc}    
\usepackage{hyperref}       
\usepackage{url}            
\usepackage{booktabs}       
\usepackage{amsfonts}       
\usepackage{nicefrac}       
\usepackage{microtype}      

\usepackage{fancyhdr}       
\usepackage{graphicx}       

\usepackage[numbers,sort&compress]{natbib} 
\usepackage{amsmath}
\usepackage{booktabs}
\usepackage{multirow} 
\usepackage{multicol}
\usepackage{caption}
\usepackage{subcaption}
\usepackage{pifont} 

\title{Teach YOLO to Remember: A Self-Distillation Approach for Continual Object Detection}

\author{
    Riccardo De Monte \\ 
    University of Padova, Italy \\ 
    \texttt{riccardo.demonte@phd.unipd.it} \\ \And 
    Davide Dalle Pezze \\ 
    University of Padova, Italy \\ 
    \texttt{davide.dallepezze@unipd.it} \\ \And
    Gian Antonio Susto \\ 
    University of Padova, Italy \\ 
    \texttt{gianantonio.susto@unipd.it} \\
}

\begin{document}
\maketitle

\begin{abstract}
Real-time object detectors like YOLO achieve exceptional performance when trained on large datasets for multiple epochs. However, in real-world scenarios where data arrives incrementally, neural networks suffer from catastrophic forgetting, leading to a loss of previously learned knowledge. To address this, prior research has explored strategies for Class Incremental Learning (CIL) in Continual Learning for Object Detection (CLOD), with most approaches focusing on two-stage object detectors. However, existing work suggests that Learning without Forgetting (LwF) may be ineffective for one-stage anchor-free detectors like YOLO due to noisy regression outputs, which risk transferring corrupted knowledge.
In this work, we introduce YOLO LwF, a self-distillation approach tailored for YOLO-based continual object detection. 
We demonstrate that when coupled with a replay memory, YOLO LwF significantly mitigates forgetting. Compared to previous approaches, it achieves state-of-the-art performance, improving mAP by +2.1\% and +2.9\% on the VOC and COCO benchmarks, respectively.
\end{abstract}


\section{Introduction}
Object detection is one of the most important research areas concerning computer vision, with many applications, from autonomous driving \citep{bogdoll2022anomaly} to medical applications \citep{ragab2024comprehensive}. Thanks to Deep Learning, several object detectors reach remarkable performance \citep{ren2016faster, ragab2024comprehensive, wang2024yolov10}.
Moreover, architectures such as RT-DETR \citep{ragab2024comprehensive} and YOLO \citep{redmon2016you,Jocher_Ultralytics_YOLO_2023,wang2024yolov10} reach a good trade-off between accuracy and speed, allowing real-time object detection. In particular, YOLO is nowadays a widely used CNN-based object detector in most applications.

\begin{figure*}[!ht]
   \centering
   \includegraphics[width=\linewidth, trim = 0 0 0 0]{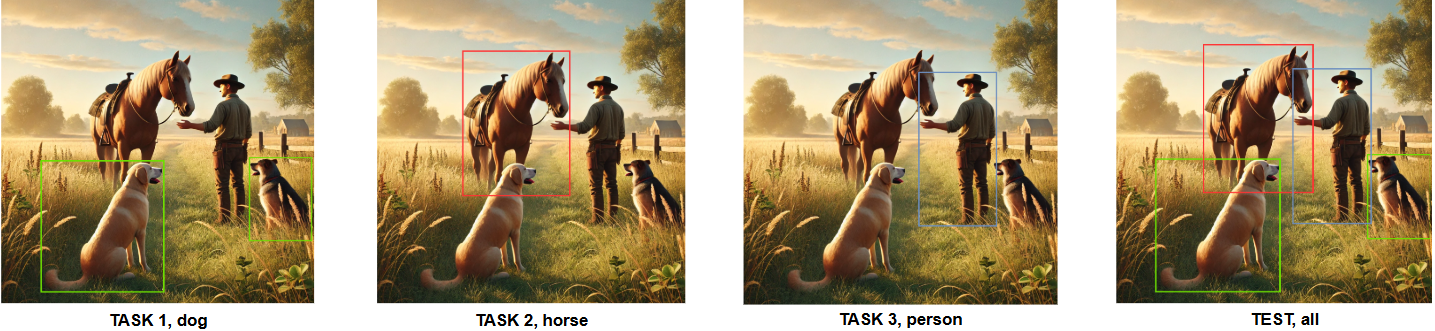}
   \caption{Artificial example for the missing annotation problem in Continual Learning for Object Detection. Objects of classes for previous tasks might be considered as background in future tasks (e.g., class "dog" in Task 2 and Task 3). In a real scenario, the three images might be different, but the same issue might arise. 
   }
   \label{fig:CLOD_scheme}
\end{figure*}

\noindent On the other hand, these object detectors are trained in the ideal scenario where all the data are given at once. Such a situation doesn't reflect real world scenarios in which models are required to learn incrementally, acquiring knowledge over time. For this purpose, Continual Learning (CL) methods aim to promote new knowledge acquisition and avoid the so-called catastrophic forgetting (CF) \citep{mccloskey1989catastrophic}, namely the tendency of neural networks to lose previously learned knowledge when exposed to new data.

When considering Continual Learning for Object Detection (CLOD), the most common scenario is
Class Incremental Learning (CIL), where an object detector learns from a stream of tasks, where each new task presents unseen object classes.
However, contrary to single-label image classification, in CLOD, previous instances could reappear in future tasks with a different ground truth (new classes).
This causes additional issues, such as the \textit{missing annotation problem}, for which new images miss annotations for old classes, as in Fig. \ref{fig:CLOD_scheme}. For this reason, consolidated methods like Experience Replay show poor performance, requiring adaptation to make it effective in the CLOD context \citep{de2024replay}.

\noindent CL literature for single-label image classification suggests that Learning without Forgetting (LwF) \citep{li2017learning}, a kind of self-distillation approach can be a valid approach, especially if coupled with a replay buffer as in Dark Experience Replay (DER) \citep{buzzega2020dark}. However, previous works \citep{peng2021sid, de2024replay} show that distillation-based approaches might not be effective for one-stage and anchor-free object detectors as YOLO.

\noindent Therefore, in our work, we rethink LwF in the context of the YOLO object detector, achieving the best performance for all CLOD benchmarks proposed in the literature and therefore showing the effectiveness of self-distillation also in the case of YOLO, especially if coupled with a replay buffer. 

Therefore, we make the following contributions:
  \begin{itemize}
        \item We propose an adaptation of LwF for YOLO-based architectures, and we call it YOLO LwF.
        \item We conduct extensive experiments on YOLOv8 architecture for VOC and COCO benchmarks, demonstrating how YOLO LwF can be a valid CL strategy. 
        \item By coupling YOLO LwF with a replay memory, our approach outperforms all competing methods, demonstrating superior performance and improving mAP by +2.1\% and +2.9\% on the VOC and COCO benchmarks, respectively.
    \end{itemize}

\section{Related Work}

Most of the CLOD literature focuses on two-stage object detectors such as Faster R-CNN \citep{peng2020faster, cermelli2022modeling, menezes2023continual, liu2023augmented, mo2024bridge}. All these works propose some solutions to prevent forgetting by leveraging the different parts of the Faster R-CNN architecture. Faster ILOD \citep{peng2020faster} proposes a distillation loss for each of the three components of the network, namely the feature extractor, the Region Proposal Network (RPN), and the class-level classification and bounding box regression network (RCN). \cite{cermelli2022modeling} revisits Faster ILOD by modifying both the distillation loss for RCN and the usual loss of Faster R-CNN in order to solve the missing annotation problem. Instead, ABR \citep{liu2023augmented} solves the missing annotation problem by saving box images from previous tasks and adding them to current images with both MixUp and Mosaic augmentation. Additionally, they replace the RPN distillation loss introduced in \cite{peng2020faster} with the so-called Attentive RoI Distillation (ARD). \cite{mo2024bridge} proposes to train an additional model on the current task to then jointly distill the past knowledge of the old teacher and the new knowledge from the new teacher.

However, nowadays, the trend in object detection is to design fast object detectors by allowing real-time detection.
For this reason, other works focus on one-stage object detectors. In \cite{peng2021sid}, the authors highlight the problem of the noisy regression output for one-stage object detectors. They propose a distillation-based method, called SID, for the FCOS architecture \citep{tian2019fcos}. In particular, they propose to distill the classification output and from intermediate features. Additionally, they propose an Inter-Related Distillation loss to maintain the inter-relation between features from different training instances.
\cite{feng2022overcoming} introduce Elastic Response Distillation (ERD) for GFLV1 \citep{li2020generalized} and FCOS \citep{tian2019fcos}. ERD performs an L2 distillation for the classification head and a KL divergence for the regression one. To overcome the noisy output problem highlighted in SID, ERD selects the candidate bounding boxes both at the classification head and regression one. This is done by applying the so-called Elastic Response Selection (ERS).
\cite{de2024replay} proposes a method called RCLPOD for YOLOv8 \citep{Jocher_Ultralytics_YOLO_2023} architecture. RCLPOD doesn't rely on output distillation, instead, it exploits Label Propagation combined with a balanced replay buffer to mitigate the forgetting. Additionally, they rely on intermediate feature distillation to improve stability.

Following RCLPOD, in this work, we focus on the YOLO object detector, but contrary to previous works on one-stage object detectors, we propose a LwF-based approach exploiting fully the network output, namely also the regression one.

\section{Methodology}

\subsection{Preliminaries}

\noindent Here, we provide a brief description of the main futures of modern YOLO detectors, justifying the need to adapt LwF in this context. In particular, we describe how the YOLO output is processed during both training and inference to highlight the reason why LwF should not be applied in a naive manner. Since in our experiments we test YOLOv8, as done for RCLPOD, we focus on it, but most of the features described are shared among modern versions of YOLO.

\medskip

Recent versions of YOLO are one-stage and anchor-free object detectors, namely, neither any region proposal nor any prior bounding box is required. From YOLOv8 to the newest versions, the loss is made of three components: the Binary Cross-Entropy (BCE) loss for classification, the Complete Intersection over Union (CIoU) loss \citep{zheng2021enhancing} and the Distance Focal Loss (DFL) \citep{li2020generalized} for regression.

\medskip

\begin{figure}[h!]
\centering
\begin{subfigure}[t]{.5\textwidth}
  \centering
  \includegraphics[width=.54\linewidth]{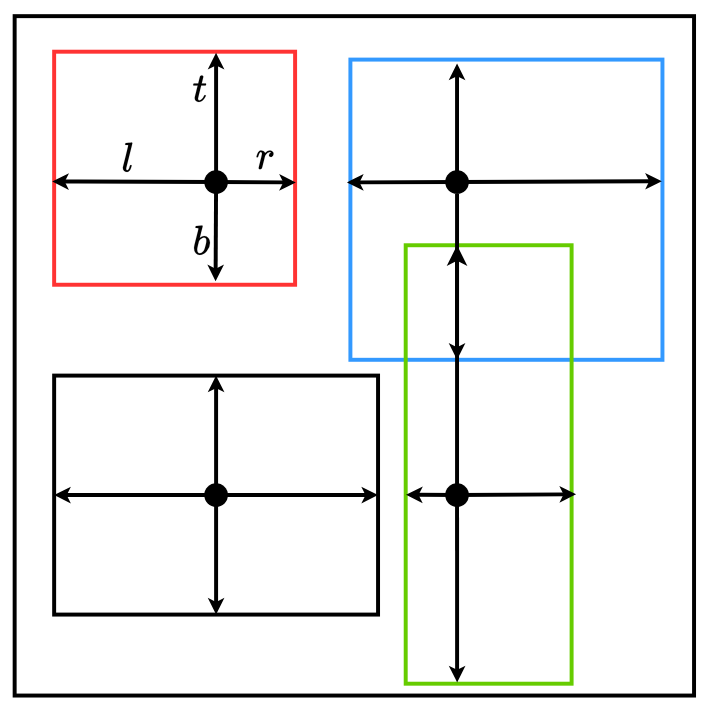}
  \caption{Example of YOLO output for $s=2$. For each anchor \\point (black dot),
  the $4\cdot L$ values are processed to\\ compute $(l,t,r,b)$, namely the four offsets.}
  \label{fig:yolov8_output}
\end{subfigure}%
\begin{subfigure}[t]{.5\textwidth}
  \centering
  \includegraphics[width=.72\linewidth]{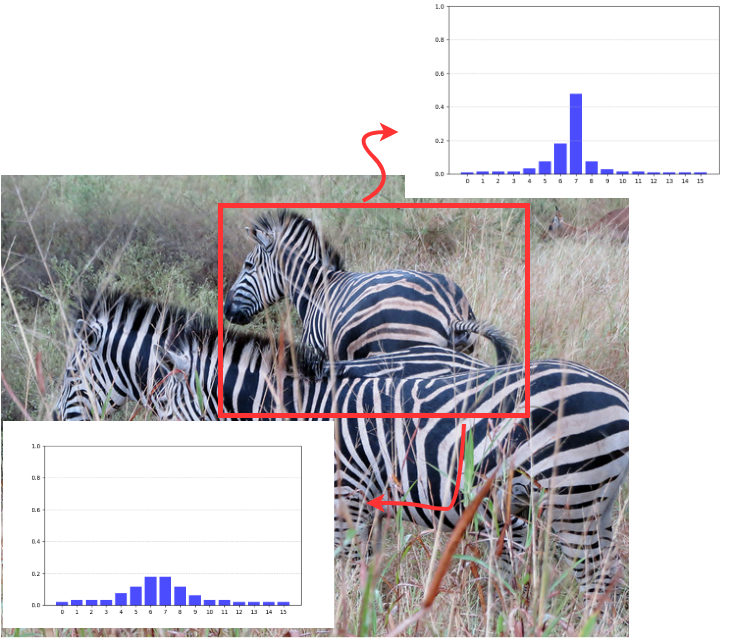}
  \caption{Occlusion example: due to occlusion, the model is more confident for the top offset and less for the bottom one.}
  \label{fig:dfl}
\end{subfigure}
\caption{Relevant YOLO features for CLOD.}
\label{fig:yolostuffs}
\end{figure}

To predict objects at different scales, YOLOv8, like all modern versions of YOLO, has three different heads, one per scale. Given an input image $\boldsymbol{x}\in\mathbb
R^{3\times d\times d}$, YOLO outputs three feature maps $\boldsymbol{F}_i\in\mathbb{R}^{(N_c+4\cdot L)\times s_i\times s_i}\;\;\;i=1,2,3$, where $N_c$ is the number of classes, $L$ is an hyper-parameter for regression and $s_i$ is such that $s_1=20$, $s_2=40$ and $s_3=80$. From now on, to simplify the discussion, we just assume that YOLO returns one single feature map $\boldsymbol{F}\in\mathbb{R}^{(N_c+4\cdot L)\times s\times s}$. Each of the $s^2$ vectors of length $N_c+4\cdot L$ corresponds to one predicted bounding box, and each of them is associated with one of the $s^2$ anchor points. For instance, Fig. \ref{fig:yolov8_output} presents a simplified example in which $s=2$, namely, the network predicts just 4 bounding boxes. For any anchor point, the first $N_c$ values of the corresponding predicted vector are used as input to sigmoid for the computation of classification scores (as in any multi-label problem), while the remaining $4\cdot L$ ones are processed as follows:

\begin{enumerate}
    \item This vector is divided into four chunks of dimension $L$, and each chunk is used for the computation of the offsets from the anchor point to the four sides of the bounding box, as in Fig. \ref{fig:yolov8_output}. 
    \item Since YOLOv8, like YOLOv10 and YOLOv11, relies on the DFL for regression, for each offset, YOLO returns a categorical distribution over $L$ possible offset values, as in Fig. \ref{fig:dfl}. For more details,  refer to \cite{li2020generalized}.
    \item The actual offset, used both during inference and for the CIoU loss computation, is just derived by computing the softmax and the corresponding expected value. 
\end{enumerate}

\noindent While during training, supervision allows to discard bad predictions \citep{feng2021tood}, during inference, those bounding boxes that have all classification scores below $0.5$ (or below any designed threshold) are discarded \footnote{In case of YOLOv8, Non-Maximum Suppression (NMS) is also used.}. This post-processing is needed since YOLO, independently by the input image, predicts $s_i^2$ bounding boxes at each scale level. Therefore, given one input image, YOLOv8 predicts $20^2+40^2+80^2=8,400$ bounding boxes. 
From the point of view of LwF, as already highlighted in previous works, most of the regression outputs are just noise since, ideally, after post-processing, most of them would have been discarded. For this reason, applying LwF with L2 loss can cause a loss of plasticity due to the meaningless supervision of the teacher.
\bigskip

\subsection{YOLO LwF}

\noindent For the reasons highlighted in the previous section, we propose some modifications to LwF to address the problem of noisy regression predictions.

\medskip

\subsubsection{Cross Entropy for the Regression Output}

\noindent Since we are in a Multi-Label setting, concerning the classification output, there is no particular reason to prefer a Cross-Entropy (CE) loss instead of the simple Mean-Squared Error (MSE) between the logits. However, for regression, by recalling that for a given anchor point and one of the four off-sets, YOLOv8 predicts a distribution over $L$ possible distance values, the choice between the CE loss or the MSE might matter. We conjecture that in the presence of noisy output provided by the teacher, matching the logits could be an additional constraint that prevents plasticity. Moreover, we think that, accordingly to \cite{kim2021comparing}, by matching logits, the teacher might transfer more corrupted knowledge, causing in some cases more forgetting. For this reason, as done in ERD \citep{feng2022overcoming}, we propose the use of CE with temperature $\tau=2.0$ for the regression output. 

\noindent Formally, given any input image $\boldsymbol{x}$, we denote with $\hat{Q}^{\text{s}}_{k,i}(\boldsymbol{x})$ the output distribution over $L$ values for anchor point $k=1,\dots, s^2$ and off-set $i=1,\dots,4$ , for the student, and with $\hat{P}^{\text{t}}_{k,i}(\boldsymbol{x})$ for the teacher counterpart. The LwF regression loss we propose is:

\begin{equation}\label{eq:reg}
    \mathcal{L}_{\text{LwF-reg}}(\boldsymbol{x})=\frac{1}{s^2\cdot 4}\sum_{k=1}^{s^2}\sum_{i=1}^4H\left(\hat{P}^{\text{t}}_{k,i}(\boldsymbol{x}),\,\hat{Q}^{\text{s}}_{k,i}(\boldsymbol{x}) \right)
\end{equation}

By applying the cross entropy between the regression output and the teacher one, the model is encouraged to focus on learning the overall distribution of possible bounding box offsets rather than focusing only on keeping the same logits.

\subsubsection{Prediction-wise Weighted Cross Entropy}
However, the issue of potentially noisy predictions from the teacher remains.
To address this, we propose to perform a prediction-wise weighted cross-entropy loss function (WCE).
Based on the specific sample and the produced predictions during training, we provide a specific weight in order to limit the harmful, noisy predictions and give more attention to the predictions that are more reliable. Contrary to ERD \citep{feng2022overcoming}, we think that exploiting the classification confidence is more reliable than confidence derived from the regression output itself.

To perform this, we weight each CE by the confidence of the teacher that any object is present, which is commonly referred to as the \textit{objectiveness score}.
Contrary to previous versions of YOLO, this score is not given by the network but can be easily derived by just computing the maximum classification score among $N_c$ scores. Formally, we modify eq. \ref{eq:reg} as follows:

\begin{align}\label{eq:weight}
    w_k\;=\;&\;\max_{j\in\{1,\dots, N_c\}}\hat{P}^{\text{t}}_{k}(\boldsymbol{x})_j\\
    \mathcal{L}_{\text{LwF-reg}}(\boldsymbol{x})\;=\;&\;\frac{1}{s^2\cdot 4}\sum_{k=1}^{s^2}\sum_{i=1}^4w_k\cdot H\left(\hat{P}^{\text{t}}_{k,i}(\boldsymbol{x}),\,\hat{Q}^{\text{s}}_{k,i}(\boldsymbol{x}) \right)
\end{align}

\begin{figure}[h!]
\centering
\begin{subfigure}{.5\textwidth}
  \centering
  \includegraphics[width=.45\linewidth]{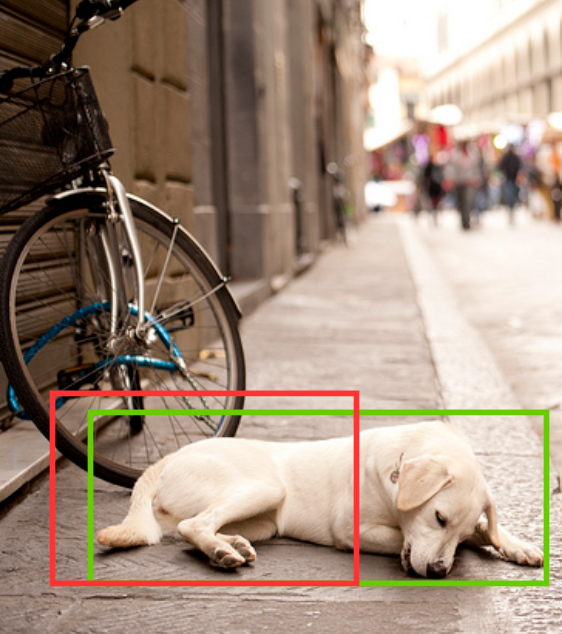}
  \caption{Higher overlapping case.}
  \label{fig:cls-sub1}
\end{subfigure}%
\begin{subfigure}{.5\textwidth}
  \centering
  \includegraphics[width=.45\linewidth]{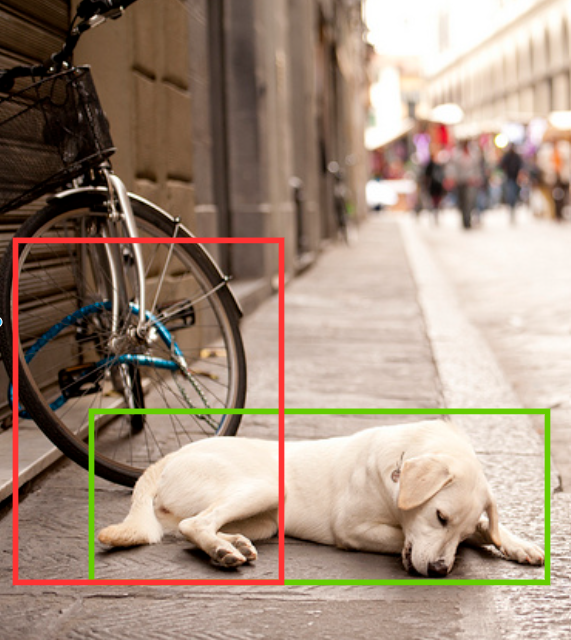}
  \caption{Lower overlapping case.}
  \label{fig:cls-sub2}
\end{subfigure}
\caption{For the same anchor point, in red is the predicted bounding box from the student, while in green is the teacher one. In \ref{fig:cls-sub1}, the student should match the teacher's output. In \ref{fig:cls-sub2}, the teacher might assign the label "dog", while the student should assign both "bike" and "dog" labels, with less confidence.}
\label{fig:cls}
\end{figure}

\noindent where $\hat{P}^{\text{t}}_{k}(\boldsymbol{x})_j$ denotes the classification score for prediction $k$ and class $j$, once image $\boldsymbol{x}$ is given as input to the teacher network. Regarding classification, we additionally propose to weigh the classification loss between the student and the teacher by the level of accordance between the corresponding bounding boxes. In fact, the classification score should coincide just in case the two bounding boxes concern the same location area. See Fig. \ref{fig:cls} for two opposite cases. Therefore, inspired by \cite{feng2021tood}, we compute the IoU for each pair of bounding boxes, $(\hat{b}_k^\text{s},\hat{b}_k^\text{t})$ $k=1,\dots, s^2$ and we weight each classification loss as follows:

\begin{align}\label{eq:IoU}
    v_k\;=\;&\;1-\text{IoU}\left(\hat{b}_k^\text{s},\hat{b}_k^\text{t}\right)\\
    \mathcal{L}_{\text{LwF-cls}}(\boldsymbol{x})\;=\;&\;\frac{1}{s^2\cdot N_c}\sum_{k=1}^{s^2}\sum_{j=1}^{N_c}v_k\cdot \text{BCE}\left(z_{k}^{\text{t}}(\boldsymbol{x})_j,\,(z_{k}^{\text{s}}(\boldsymbol{x})_j\right)
\end{align}

\noindent where $z_{k}^{\text{t}}(\boldsymbol{x})_j$ and $z_{k}^{\text{s}}(\boldsymbol{x})_j$ denotes the logits for the anchor point $k$, class $j$, for the teacher and the student respectively.

Therefore, like for the regression loss, also for classification loss, we perform a prediction-wise weighting that allows the model to focus better on what is really relevant and avoid irrelevant information.

\subsection{Masking overlapping objects}

\medskip Finally, according to previous works \citep{de2024replay}, it is reasonable to mask old classes for the classification loss computation when labels for the current task are provided. The overall loss is then computed as follows:

\begin{equation}
    \mathcal{L}(\boldsymbol{x})=\mathcal{L}_{\text{YOLO-mask}}(\boldsymbol{x})+\beta_1\mathcal{L}_{\text{LwF-reg}}(\boldsymbol{x})+\beta_2\mathcal{L}_{\text{LwF-cls}}(\boldsymbol{x})
\end{equation}
\noindent where $\mathcal{L}_{\text{YOLO-mask}}(\boldsymbol{x})$ is the usual YOLO loss for which masking is applied and $\beta_1,\beta_2$ are two constants used to control the trade-off plasticity-stability.

\subsection{Combine self-distillation with experience replay}

\noindent As highlighted in most of the literature on CL for single-label classification, Experience Replay (ER) is a really effective approach. Moreover, as shown in RCLPOD \citep{de2024replay}, a replay memory is necessary for realistic scenarios in which the stream of tasks is quite long. For this reason, we propose to exploit a replay buffer to store samples from the old task and apply YOLO LwF to both samples from the buffer and the new ones. Contrary to single-label scenarios, in Object Detection we cannot naively exploit the labels for old samples since task interference can arise. There are several possible ways to solve this issue: as done in RCLPOD, update the memory by adding pseudo-labels, keep a mask for each sample to identify the classes of interest, or simply ignore labels and rely only on self-distillation. We proceed with this last option to both reduce the complexity of our method and, at the same time, avoid low-quality labels that the model might produce if pseudo-labels are added.

\noindent As pointed out in RCLPOD, in a Multi-label CIL setting, balancing the number of examples per class is quite effective in keeping a balanced accuracy over different classes. For this reason, instead of a random update for the replay buffer, we apply Optimizing Class Distribution in Memory (OCDM) \citep{liang2021optimizing}, as already proposed in RCLPOD. This method applies a greedy approach to keep just the samples from both the memory and the new data to target a uniform data distribution over classes.

\noindent Another possible approach is to follow what has been proposed in DER \citep{buzzega2020dark} for single-label classification, where for each sample, the corresponding output network is saved in the replay buffer. However, in addition to the task interference problem that can be solved by masking, in the case of YOLO saving the entire output of YOLO for each sample in memory is not practicable: recalling that YOLOv8 predicts 8,400 bounding boxes, and for each of them we have a $(N_c+4\cdot L)$ vector and in the simple case in which $N_c=1$ and $L=16$ (default value), more than 2MB are needed to store the output for one single image.
For example, saving 800 samples in the replay memory for COCO requires more than 1,6 GB  of additional storage (excluding the stored images).
For this reason, we stick with the simple combination of LwF with OCDM for the memory update as the most efficient and sustainable solution.

\section{Experimental Setting}\label{sec:experimental_setting}
\medskip
\subsection{Evaluation Protocol and Metrics}
\noindent As done in previous works~\citep{shmelkov2017incremental, peng2021sid, menezes2023continual}, we test our proposed LwF-based method on the 2017 version of the PASCAL VOC~\citep{everingham2015pascal} benchmark with 20 object classes and on the Microsoft COCO challenge dataset~\citep{lin2014microsoft} with 80 object classes. 
The CIL scenarios derived by the two datasets are the same ones proposed for RCLPOD, accordingly also to other previous works \cite{shmelkov2017incremental, peng2021sid}. For clarification, we report here the same notation used for RCLPOD to distinguish the different scenarios:
$N\text{p}M$, means that the first task ($i = 1$) consists of the first $N$ classes in the list (e.g., in alphabetical order), while each subsequent task ($i > 1$) consists of the classes from $N + (i-2) \cdot M$ to $N + (i-1) \cdot M - 1$.
The CL scenarios for VOC are $15\text{p}1$, $15\text{p}5$, $10\text{p}10$ and $19\text{p}1$, while for COCO $40\text{p}40$ and $40\text{p}10$.

\noindent As in~\cite{shmelkov2017incremental, de2024replay}, to evaluate the performance of YOLOv8, we report the mean average precision (mAP) at the end of the training: with a 0.5 IoU threshold, denoted by mAP$^{50}$, for VOC, and the mAP weighted across different IoU thresholds (from 0.5 to 0.95), denoted with mAP$^{50-95}$, for COCO. 

\subsection{YOLO training details}

\noindent We test YOLO LwF on YOLOv8n (3.2M parameters), the smallest version among the ones available for YOLOv8. By doing so, we test our method in the more challenging scenario, where the capacity of the model is the lowest possible. We initialize the backbone
parameters with the ones pre-trained on Image-Net, available on \cite{Jocher_Ultralytics_YOLO_2023} as done for RCLPOD and in CLOD literature \citep{menezes2023continual}.
\noindent We follow the training procedure suggested in \cite{Jocher_Ultralytics_YOLO_2023,wang2024yolov10} and we report the training hyper-parameters in Table \ref{tab:hyper}. In particular, as done for RCLPOD, we set the number of epochs for each task to 100.

\subsection{CL baselines}

\noindent Firstly, we report the results for the ideal Joint Training, and for Fine-Tuning, a kind of lower bound for what concerns stability but still a good baseline for plasticity.
Since our goal is to improve LwF, we report the results also for LwF, and, for a fair comparison with our method combined with OCDM, we test the combination of LwF with OCDM (LwF+OCDM). 
Additionally, we also report the results for RCLPOD \citep{de2024replay}, the current SOTA for one-stage CLOD.

\noindent As for RCLPOD, if a replay buffer is used, the capacity is fixed to around 5\% of the entire dataset, namely 800 images for PASCAL VOC and 6,000 for COCO.

\noindent For LwF, by following \cite{shmelkov2017incremental}, we set the distillation loss weight $\lambda$ to 1.
For Pseudo-Labeling (used by RCLPOD), to ensure consistency with inference, 
as in the original work, we set the classification threshold to $0.5$ and the IoU threshold, for Non-maximum Suppression, to $0.7$.

Even if ERD \citep{feng2022overcoming} is originally proposed for GFLV1 (with ResNet50 as backbone), we implement ERD for YOLOv8, and, contrary to the original work, we test it also on VOC. As in the original work, we set the hyper-parameters $\alpha_1,\alpha_2$ to 2 and $\lambda_2$ to 1. Contrary to the original work, we set $\lambda_1=1\cdot10^{-2}$. As done for LwF, we also ERD with a replay buffer with OCDM update, and we denote it with ERD+OCDM.

\noindent Finally, for YOLO LwF we set $\beta_1=4\cdot 10^3$ and $\beta_2=0.5\cdot 10^3$ for all experiments. These values work well independently in the CL scenario.

\section{Results}

\noindent In this section, we present the resulting performance of our method compared to the baselines presented in the previous section. Tab. \ref{tab:results} presents the final mAP obtained at the end of the last task for each CLOD scenario. Tab. \ref{tab:lwfvs} compares the methods by the stability-plasticity perspective, regarding the two longest CLOD scenarios, namely VOC15p1 and COCO 40p10, while Tab. \ref{tab:oldnew}
for 2-tasks scenarios. In Sec.\ref{sec:res-short} and \ref{sec:res-long} we discussed the results obtained, while in Sec. \ref{sec:ablation} we present an ablation study that justifies our proposed method.

\begin{figure}[h!]
\centering
\begin{subfigure}{.5\textwidth}
  \centering
  \includegraphics[width=1.0\linewidth]{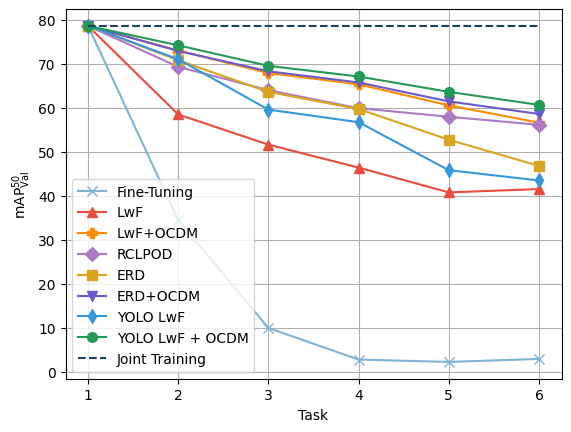}
  \caption{VOC15p1 scenario, mAP$^{50}$}
  \label{fig:coco15p1}
\end{subfigure}%
\begin{subfigure}{.5\textwidth}
  \centering
  \includegraphics[width=1.0\linewidth]{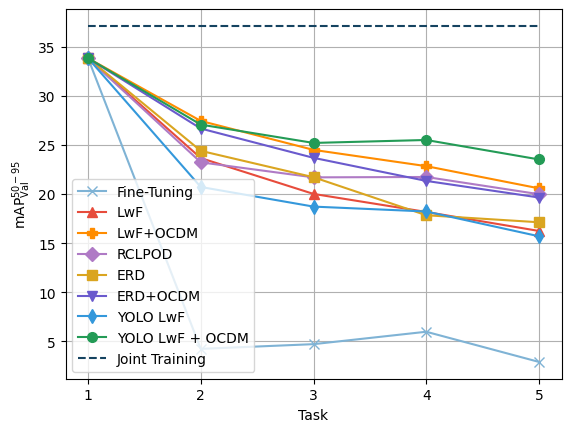}
  \caption{COCO 40p10 scenario, mAP$^{50-95}$}
  \label{fig:coco40p10}
\end{subfigure}
\caption{Results for the two long scenarios: VOC 15p1 and COCO 40p10.}
\label{fig:results-plot}
\end{figure}

\subsection{Two-task scenarios} \label{sec:res-short}
When no replay buffer is used, Tab. \ref{tab:results} shows that YOLO LwF outperforms the other methods in all the two-task scenarios, except for the VOC19p1, where ERD is the best one. In particular, in the VOC 10p10 and COCO 40p40 scenarios, our method shows strong performance even without the use of any replay buffer. This is justified by the fact that these two scenarios are such that the second task also presents instances of classes from the previous task. 
In all four scenarios, YOLO LwF improves the results of the standard LwF implementation. In particular, Tab. \ref{tab:lwfvs} shows that our approach benefits from higher plasticity and, at the same time, in three of four scenarios, less forgetting is observed. The only scenario where ERD outperforms our proposed method is such that the number of new classes is much lower than the ones in the previous one. In fact, Tab. \ref{tab:lwfvs} shows that this gap is mainly due to the ERD property of suffering less forgetting by reducing plasticity. Therefore, for what concerns distillation-based methods, we can state that our method is the best in balancing stability and plasticity in the two-task scenarios.

\begin{table*}[th]
\caption{Results for YOLOv8n. Each column represents one of the studied scenarios based on VOC and COCO datasets. Each row represents a different tested CL technique. In bold is the best method and underlined is the second best method. As in CLOD literature, mAP50 is used for VOC-based scenarios and mAP 50-95 for COCO-based scenarios. }
\centering
\begin{tabular}{lcccc|cc}
\toprule
\multirow{2}{*}{\textbf{CL Method}}  
& \multicolumn{4}{c}{\textbf{VOC (mAP$^{50}$)}} & \multicolumn{2}{|c}{\textbf{COCO (mAP$^{50-95}$)}}       \\ 
        &  \textbf{10p10} &  {\textbf{19p1}} &  {\textbf{15p5}} &  {\textbf{15p1}} &  {\textbf{40p40}} & \textbf{40p10} \\
        \midrule
Joint Training    &  {78.5}          &  {78.5}         &  {78.5}              &  {78.5}         &  {37.3}          & 37.3          \\  
Fine-Tuning       &  {36.4}              &  {16.5}              &  {16.4}              &  {3.0}              &  {13.6}               &        2.9        \\ 
LwF\cite{}               &  {57.3}               &  {60.0}              &  {57.0}              &  {41.5}              &  {19.0}               &          16.2    \\  
LwF + OCDM&69.7&\underline{73.4}&\underline{68.9}&{56.6}&22.7&\underline{20.6}\\

RCLPOD    &  {72.5}      &  {68.4}     &  {67.7}              &  {56.1}     &  {24.1}      &   20.0           \\ 
ERD&68.6&67.0&65.3&46.8&21.9&17.1\\
ERD + OCDM&69.0&72.8&67.2&\underline{58.6}&22.2&19.6\\
\midrule
 \textbf{YOLO LwF (Ours)}&\textbf{73.1}&65.6&68.1&43.5&\underline{24.6}&15.7\\
\textbf{YOLO LwF + OCDM (Ours)}&\underline{72.9}&\textbf{73.7}&\textbf{71.0}&\textbf{60.7}&\textbf{27.3}&\textbf{23.5}\\
\bottomrule
\end{tabular}

\label{tab:results}
\end{table*}

Concerning the use of a replay buffer, our method reaches the lowest forgetting in all scenarios, except for COCO40p40. In comparison to LwF with OCDM and ERD with OCDM, YOLO LwF with OCDM doesn't suffer from a lack of plasticity and has less forgetting than RCLPOD with a small gap in terms of plasticity in the VOC scenarios. 
Even if naive LwF with replay buffer is the most stable method in the COCO 40p40 scenario, YOLO LwF with OCDM is quite close and at the same time exhibits the best stability/plasticity trade-off among the other methods.

\subsection{Scenarios VOC 15p1 and COCO 40p10}\label{sec:res-long}

\noindent As shown in Fig. \ref{fig:results-plot}, the use of a replay buffer plays an important role in these two longer scenarios. In fact, these two scenarios are quite relevant for those applications in which the data from the new task does not present any example of an object from previous tasks.

In Tab. \ref{tab:oldnew}, we report the average performance in terms of forgetting (old) and plasticity (new). In particular, at each new task, we measure the mAP for old classes (from previous tasks) and the mAP for new classes, and we report in Tab. \ref{tab:oldnew} the average over all tasks.

In case no replay buffer is used, our method is the best in terms of plasticity among the distillation-based methods, as shown in Tab. \ref{tab:oldnew}. However, as for two-task scenarios, ERD shows stronger performance thanks to its resistance to forgetting. As anticipated for the shorter scenarios, we think that since our method exploits the classification confidence to properly weight the more reliable teacher predictions, in long scenarios where fewer instances of old classes might appear, our method is prone to forget due to the low amount of meaningful knowledge from the teacher. This fact is confirmed when a replay buffer is additionally used.

When distillation-based approaches are coupled with a replay buffer,  our proposed approach outperforms all the others in both scenarios, showing the adaptability of our approach when combined with a replay memory. Regarding COCO 40p10, contrary to ERD, our method improves in terms of stability (from 16.8 to 25.1) with a slight decrease of plasticity (from 27.8 to 27.5). Concerning VOC 15p1, our approach still stands out thanks to a remarkable resistance to forgetting. However, as with all distillation-based methods, there is still an important gap in terms of plasticity with respect to RCLPOD. This fact highlights that there is still room for improvement of the distillation-based approaches to close the gap with Joint-Training.

\begin{table*}[!htb]
\centering
\caption{Results for YOLOv8n for for 2-tasks scenarios. "old" is the mAP for the classes of Task 1 after Task 2 training, "new" is the mAP for the new classes, while "all" is the mAP over all the classes.}
  \begin{tabular}{lccc|ccc|ccc|ccc}
    \toprule
    \multirow{2}{*}{\textbf{CL Method}} &
      \multicolumn{3}{c}{\textbf{VOC 10p10}}&\multicolumn{3}{c}{\textbf{VOC 19p1}}&\multicolumn{3}{c}{\textbf{VOC 15p5}}&\multicolumn{3}{c}{\textbf{COCO 40p40}}\\
      & {old} & {new} & {all}& {old} & {new} & {all}& {old} & {new} & {all}&old&new&all\\
      \midrule
          Fine-Tuning & 0.0 & 72.8 &36.4&14.0 & \textbf{64.8} & 16.5&3.8&54.2& 16.4 & 0.0&\textbf{27.1}&13.6\\
    LwF & 50.7 & 63.8 & 57.3 &60.6&50.3& 60.0&61.6&43.4& 57.0 &26.9&11.0&19.0\\
    LwF + OCDM & 72.5 & 66.9 & 69.7 &\underline{74.3}& 56.4&\underline{73.4}&\underline{76.1}&47.3&\underline{68.9}& \textbf{31.8}&13.7&22.7 \\
    RCLPOD&70.9&\textbf{74.1}&72.5&68.8&61.8&68.4&71.6&\textbf{55.8}&67.7&25.7&22.5&24.1\\
    ERD&71.1&66.2&68.6&67.7&53.5&67.0&72.7&43.1&65.3&28.6&15.1&21.9\\
    ERD + OCDM&\underline{72.7}&65.3&69.0&73.8&54.6&72.8&75.3&43.0&67.2&\underline{30.8}&13.5&22.2\\
    \midrule
    \textbf{YOLO LwF (Ours)} & \underline{72.7}&\underline{73.4}&\textbf{73.1}&65.8 &\underline{62.1}&65.6&72.8&54.3&68.1& 23.8&\underline{25.5}&\underline{24.6}\\
    \textbf{YOLO LwF + OCDM (Ours)}&\textbf{73.5}&72.3&\underline{72.9}&\textbf{74.4}&60.7&\textbf{73.7}&\textbf{76.5}&54.4&\textbf{71.0}&29.4&25.2&\textbf{27.3}\\
    \bottomrule
  \end{tabular}
  
  \label{tab:lwfvs}
\end{table*}

\begin{table*}[!htb]
  \caption{Results for YOLOv8n for VOC 15p1 and COCO 40p10 scenarios. "old" is the average mAP for the old classes over all the tasks, "new" is the average mAP for the new classes over all the tasks.}
\centering
  \begin{tabular}{lccccccccc}
    \toprule
    \multirow{2}{*}{\bf CL Method} &&
      \multicolumn{3}{c}{{\bf VOC 15p1} ({\bf mAP}$^{50}$)}&&\multicolumn{4}{c}{{\bf COCO 40p10} ({\bf mAP}$^{50-95}$)}\\
      && {old} & {new} &all&& {old} & {new}&{all}&\\
      \toprule
      Fine-Tuning&&9.0&38.3&3.0&&0.0&{\bf28.7}&2.9\\
      LwF&&48.8&31.6&41.5&&20.5&15.4&16.2\\
      LwF + OCDM&&66.5&34.8&56.6&&\textbf{25.2}&17.6&\underline{20.6}\\
    RCLPOD && 62.6&\textbf{43.0} &56.1&&  21.6&22.6&20.0\\
        ERD&&60.4&32.2&46.8&&20.7&19.2&17.1\\
    ERD + OCDM&&\underline{67.4}&33.2&\underline{58.6}&&24.0&17.4&19.6\\
    \midrule
    \textbf{YOLO LwF (Ours)}&&56.4&\underline{39.0}&43.5&&16.8&\underline{27.8}&15.7\\
    \textbf{YOLO LwF + OCDM (Ours)}&&\textbf{69.0}&34.2&\textbf{60.7}&&\underline{25.1}&27.5&\textbf{23.5}\\
    \bottomrule
  \end{tabular}

  \label{tab:oldnew}
\end{table*}

\subsection{Ablation}\label{sec:ablation}

Here, we discuss the various techniques proposed to adapt LwF for YOLO. Regarding the weights proposed in equation \ref{eq:IoU}, we refer to it with the notation Cls-IoU.

In Tab. \ref{tab:10p10}, we report the mAP increment obtained by adding each component progressively in the VOC 10p10 scenario. The use of CE instead of L2 allows a significant improvement both in stability and plasticity, with a resulting increment of 14.5 mAP. By using WCE, as in equation \ref{eq:weight}, we get an additional increment for both plasticity and stability.
By masking old classes when new labels are given, we get a stability improvement with a slight reduction of plasticity. Finally, Cls-IoU recovers the plasticity with a small degradation in stability.

\noindent We additionally investigate the role of WCE and Cls-IoU in the case of YOLO LwF with a replay buffer in the challenging scenario VOC 15p1. As reported in Tab. \ref{tab:15p1}, WCE is even more important for longer scenarios, with a 2.7 mAP increment. Additionally, Cls-IoU allows a significant plasticity improvement without sacrificing too much stability.
Therefore, this ablation study shows the importance of CE and highlights the effectiveness of WCE and Cls-IoU for a better stability-plasticity trade-off.
\begin{table*}[!htb]
\begin{minipage}[t]{0.45\textwidth}
\centering
\caption{Ablation study on VOC 10p10 scenario. No replay buffer is used.}
  \begin{tabular}{c|c|c|c|ccc}
    \toprule
    \multirow{2}{*}{CE}&\multirow{2}{*}{WCE}&\multirow{2}{*}{mask}&\multirow{2}{*}{Cls-IoU}&\multicolumn{3}{c}{\textbf{VOC 10p10 (mAP$^{50}$)}}\\
    
    &&&&old&new&all\\
    \midrule
    \ding{55}&\ding{55}&\ding{55}&\ding{55}&50.7&63.8&57.3\\
    \ding{51}&\ding{55}&\ding{55}&\ding{55}&71.0&72.8&71.8\\
    \ding{51}&\ding{51}&\ding{55}&\ding{55}&71.5&73.2&72.3\\
    \ding{51}&\ding{51}&\ding{51}&\ding{55}&\textbf{72.9}&73.1&{73.0}\\
    \ding{51}&\ding{51}&\ding{51}&\ding{51}&72.7&\textbf{73.4}&\textbf{73.1}\\
    \bottomrule
  \end{tabular}
  \label{tab:10p10}
  \end{minipage}
    \hfill
    \begin{minipage}[t]{0.45\textwidth}
        \caption{Ablation study on VOC 15p1 scenario. The replay buffer is used.}
  \begin{tabular}{cccc}
    \toprule
    \multirow{2}{*}{YOLO LwF + OCDM}&\multicolumn{3}{c}{\textbf{VOC 15p1 (mAP$^{50}$)}}\\
    
    &old&new&all\\
    \midrule
    w/o WCE&66.7&31.2&57.3\\
    w/o Cls-IoU&\textbf{69.2}&32.8&\textbf{61.0}\\
    w WCE,\, w Cls-IoU &69.0&\textbf{34.2}&60.7\\
    \bottomrule
  \end{tabular}
  \label{tab:15p1}
    \end{minipage}
  
  \label{tab:ablation}
\end{table*}

\section{Conclusions}

In this work, we propose a distillation-based approach for modern YOLO object detectors, called YOLO LwF. As shown in previous CLOD works, distilling knowledge from the regression output to reach a good stability-plasticity is extremely challenging. In our work, we analyze the main features of YOLO, justifying the nature of this problem. Firstly, we show the effectiveness of using a Cross-Entropy loss for regression distillation. To reduce the noisy regression problem, we propose to couple the regression distilled knowledge with the classification one, and, vice versa, to couple the classification one with regression. 

We demonstrate that when coupled with a replay memory, YOLO LwF significantly mitigates forgetting. Compared to previous approaches, it achieves state-of-the-art performance, improving mAP by +2.1\% and +2.9\% on the VOC and COCO benchmarks, respectively.

In short CLOD scenarios, YOLO LwF shows strong performance, even without the use of any additional replay memory. When combined with a replay memory, our proposed method outperforms other distillation-based approaches and the pseudo-label one RCLPOD in all benchmarks. Therefore, our work shows how distillation-based approaches can be highly effective for a real-time object detector like YOLO and defines a new baseline for future works in this field.

\newpage


\bibliographystyle{unsrtnat}

\bibliography{main}  

\newpage

\appendix
\section{Appendix}

\subsection{Training hyper-parameters}

Table \ref{tab:hyper} presents the hyper-parameters used for training YOLO. Regarding data augmentation, we follow \cite{Jocher_Ultralytics_YOLO_2023, wang2024yolov10}.

\begin{table}[h!]
  \begin{center}
    \caption{Hyper-parameters for YOLOv8.}
    \label{tab:hyper}
    \begin{tabular}{lc} 
    \hline
      \textbf{Hyper-parameter} &  \\
      \hline
      epochs & $100$\\
      optimizer & SGD \\
      batch size & $8$\\
      momentum & $0.937$ \\
      weight decay& $5\times 10^{-4}$\\
      warm-up epochs&$3$\\
      warm-up momentum& $0.8$\\
      warm-up bias learning rate & $0.1$\\
      initial learning rate & $10^{-2}$\\
      final learning rate & $10^{-4}$\\
      learning rate scheduling& linear decay\\
      box loss gain & $7.5$\\
      classification loss gain & $0.5$\\
      DFL loss gain & $1.5$\\
      $L$ (reg\_max)&$16$\\
      \hline\\
    \end{tabular}
  \end{center}
\end{table}

\subsection{ERD details}

\noindent By following the original work we set the hyper-parameters reported in Tab. \ref{tab:hyper-erd}.

\begin{table}[h!]
  \begin{center}
    \caption{ERD hyper-parameters for YOLOv8.}
    \label{tab:hyper-erd}
    \begin{tabular}{llc} 
    \hline
      \textbf{Hyper-parameter} &  \textbf{Description}&\\
      \hline
$\alpha_1,\alpha_2$ &for thresholds computation&2\\
$\lambda_1$&ERD classification gain&$0.01$\\
$\lambda_2$&ERD regression gain&$1.0$\\
$t$&temperature for ERS&$1.0$\\
$\tau$ &temperature for KL divergence&$10$\\
NMS IoU threshold &&$0.005$\\
\hline\\
    \end{tabular}
  \end{center}
\end{table}

\noindent All of them coincide with the ones proposed in the original work, except for $\lambda_1$. Based on our implementation for YOLOv8, $\lambda_1=0.01$ is a better choice as shown in Tab. \ref{tab:erd-test}

\begin{table*}[!htb]
\centering
\caption{ERD for YOLOv8 in the VOC 10p10 scenario, varying $\lambda_1$.}
\label{tab:erd-test}
\begin{tabular}{cccc}

    \toprule
    \multirow{2}{*}{$\lambda_1$}&\multicolumn{3}{c}{\textbf{VOC 10p10 (mAP$^{50}$)}}\\
    
    &old&new&all\\
    \midrule
    $1.0$&71.0&32.1&51.6\\
    $0.01$&71.1&66.2&68.6\\
    \bottomrule
  \end{tabular}
\end{table*}

\newpage
\subsection{Memory efficiency}
\label{subsec:mem}
Following RCLPOD \citep{de2024replay}, we compare our approach with RCLPOD in the VOC15p1 scenario when the memory size is reduced from 800 samples to 400 samples.
From Tab. \ref{tab:mem-size}, we can observe that by decreasing the memory size $m$  from $m=800$ (5\%) to $m=400$ (2.5\%), YOLO LwF with OCDM is much less sensitive to this hyper-parameter than RCLPOD.

\begin{table*}[!htb]
\centering
  \caption{Results in the VOC15p1 scenario varying the memory size $m$.}
  \begin{tabular}{lccc}
    \toprule  
    CL Method &
      $m=800$ & $m=400$& $\Delta\%\;(\downarrow)$\\
      \toprule
    RCLPOD& 56.1&52.5& 6.4\%\\
    YOLO LwF + OCDM (Ours) & 60.7&60.4&{\bf 0.5}\%\\
    \bottomrule
  \end{tabular}

  \label{tab:mem-size}
\end{table*}

\end{document}